\documentclass{llncs}

\usepackage[utf8]{inputenc}
\usepackage[english]{babel}
\usepackage{cite}
\usepackage{bm}                % bold italic math

\usepackage{amssymb,amsfonts,amsmath}
\usepackage{lmodern}
\usepackage{algorithmicx}
\usepackage{graphicx}
\usepackage{textcomp}
\usepackage{xcolor}
\usepackage{comment}
\usepackage{hyperref}
  \hypersetup{
    colorlinks=true,
    linkcolor=blue,
    filecolor=blue,
    urlcolor=blue,
    citecolor=blue
  }
\usepackage{multirow}
\usepackage{booktabs}
\usepackage[noabbrev,capitalise]{cleveref} % must load after other packages
  \crefname{equation}{}{}      % only the number for equations
\usepackage{tikz, upgreek, standalone}
\usepackage{fancyhdr}
\usepackage[printonlyused]{acronym}

\graphicspath{{figures/}}
\DeclareGraphicsExtensions{.pdf,.eps,.png,.jpg}

\fancypagestyle{firstpagefooter}{
  \fancyhf{}
}

\usepackage{algpseudocode}
\algnewcommand{\LineComment}[1]{\State \(\triangleright\) #1}

\begin{document}

% Title and author/institute blocks for LLNCS
\title{A Comparative Study of Feature Selection in Tsetlin Machines}

\author{%
  Vojtech Halenka\inst{1} \and
  Ole-Christoffer Granmo\inst{1} \and
  Lei Jiao\inst{1} \and
  Per-Arne Andersen\inst{1}
}

\institute{%
  Department of ICT, University of Agder, Grimstad, Norway\\
  \email{\{vojtech.halenka,ole.granmo,lei.jiao,per.andersen\}@uia.no}
}

\maketitle
\thispagestyle{firstpagefooter}
\begin{acronym}[ASIC]
    \acro{AI}{artificial intelligence}
    \acrodefindefinite{AI}{an}{an}
    \acro{ASIC}{application-specific integrated circuit}
    \acrodefindefinite{ASIC}{an}{an}
    \acro{AUC}{area under curve}
    \acro{BD}{bounded delay}
    \acro{BNN}{binarized neural network}
    \acro{CNN}{Convolutional Neural Network}
    \acro{CTM}{Convolutional Tsetlin machine}
    \acro{FSM}{finite state machine}
    \acro{FS}{Feature Selection}
    \acro{IG}{Integrated Gradients}
    \acro{HV}{hypervector}
    \acro{HVTM}{hypervector Tsetlin machine}
    \acro{LA}{learning automaton}
    \acrodefplural{LA}{learning automata}
    \acrodefindefinite{LA}{an}{a}
    \acro{ML}{machine learning}
    \acrodefindefinite{ML}{an}{a}
    \acro{NLP}{natural language processing}
    \acro{NN}{Neural Network}
    \acro{DNN}{Deep Neural Network}
    \acro{TA}{Tsetlin automaton}
    \acrodefplural{TA}{Tsetlin automata}
    \acro{TAT}{Tsetlin automaton team}
    \acro{TM}{Tsetlin machine}
    \acro{RTM}{Regression Tsetlin machine}
    \acro{ROAR}{Remove and Retrain}
    \acro{ROAD}{Remove and Debias}
    \acro{RbE}{Reasoning by Elimination}
    \acro{HVSize}{Hypervector size}
    \acro{NBits}{number of projection Booleans}
    \acro{HVTA}{Hypervector Tsetlin Automata}
\end{acronym}

\begin{abstract}
\ac{FS} is crucial for improving model interpretability, reducing complexity, and sometimes for enhancing accuracy. The recently introduced \ac{TM} offers interpretable clause-based learning, but lacks established tools for estimating feature importance. In this paper, we adapt and evaluate a range of \ac{FS} techniques for \acp{TM}, including classical filter and embedded methods as well as post-hoc explanation methods originally developed for neural networks (e.g., SHAP and LIME) and a novel family of embedded scorers derived from \ac{TM} clause weights and \ac{TA} states. We benchmark all methods across 12 datasets, using evaluation protocols, like \ac{ROAR} strategy and \ac{ROAD}, to assess causal impact. Our results show that \ac{TM}-internal scorers not only perform competitively but also exploit the interpretability of clauses to reveal interacting feature patterns. Simpler \ac{TM}-specific scorers achieve similar accuracy retention at a fraction of the computational cost. This study establishes the first comprehensive baseline for \ac{FS} in \ac{TM} and paves the way for developing specialized \ac{TM}-specific interpretability techniques.
\end{abstract}
\section{Introduction}

Machine learning models often benefit from \ac{FS} to identify the most informative input features, which can improve generalization and interpretability. In \ac{DNN}, a rich array of methods exists to assess feature importance (e.g. SHAP, LIME) \cite{Lundberg2017_SHAP,Ribeiro2016_LIME}. However, the \ac{TM} — a logic-based learning algorithm using \ac{TAT} — has not yet seen a comparable development of such interpretability-assisting tools. The \ac{TM} achieves competitive accuracy on various tasks while producing human-readable rules \cite{Granmo2020_TM}, but questions remain on how to best discern which input features drive its decisions.

\ac{FS} is particularly relevant for \acp{TM} to examine large learned rule sets and focus interpretation on key factors. While classical \ac{FS} methods and \ac{NN}-inspired explainers could potentially be applied to \acp{TM}, their efficacy in this new context is largely unexplored. Prior work has hinted that \acp{TM} may handle high-dimensional inputs without explicit \ac{FS} \cite{berge2019text,HypervectorTM}, yet a systematic comparison of \ac{FS} methods for \acp{TM} is missing.

In this paper, we bridge that gap by conducting a comparative study of \ac{FS} methods applied to \ac{TM} across a variety of benchmark datasets. We evaluate methods from all three major \ac{FS} categories—filter, embedded, and wrapper—adapting methods from both traditional statistics and modern \ac{NN} explainability. Our contributions are as follows:
\begin{itemize}
  \item We adapt and apply four post-hoc explainers to Tsetlin Machines’ binary inputs.
  \item We propose a family of \ac{TM}-internal scorers derived from clause weights and \acp{TA} states.
  \item We benchmark 23 methods on 12 UCI datasets using four evaluation protocols, tracking \ac{AUC} and ranking time.
  \item We compare speed–quality trade-offs, ranking correlations, and outline \ac{TM}-specific \ac{FS} strategies.
  \item Finally we select a few representatives from clusters of the \ac{FS} and plot them for explainability comparison.
\end{itemize}

\section{Related Work}
\label{sec:related}

\ac{FS} methods are generally categorised into:  
\begin{itemize}
  \item \textbf{Filter methods} evaluate feature relevance using data characteristics independent of any specific learning algorithm (e.g., Mutual Information \cite{Battiti1994_MI}, Chi-square \cite{Breiman2001_RF}, Variance thresholds). They are computationally efficient but ignore model interactions.
  \item \textbf{Wrapper methods} search for an optimal subset by repeatedly evaluating a predictive model. Modern wrappers include permutation importance \cite{Fisher2018_PermutationImportance} and other post-hoc probes, capturing interactions at high compute cost.
  \item \textbf{Embedded methods} perform \ac{FS} as part of model training (e.g., tree-based split importance \cite{Gini}. They balance filter speed with wrapper specificity but are model-specific.
\end{itemize}

Gradient-based and other wrapper explainers have been widely adopted in \acp{NN}, driven by the challenge of interpreting these black-box models:
\begin{itemize}
  \item \textbf{LIME} fits local linear surrogates \cite{Ribeiro2016_LIME}.
  \item \textbf{SHAP} uses Shapley values for global attributions \cite{Lundberg2017_SHAP}.
  \item \textbf{Integrated Gradients (IG)} accumulates gradients along the inputs \cite{Sundararajan2017_IG}.
  \item \textbf{SmoothGrad} and its variants average noisy‐gradient attributions \cite{Smilkov2017_SmoothGrad}.
\end{itemize}

The \ac{TM} learns human-readable propositional clauses via \ac{TAT} \cite{Granmo2020_TM}. Prior work indicates \acp{TM} scale well with high-dimensional inputs (e.g., text n-grams \cite{berge2019text}, hypervector encodings \cite{HypervectorTM}), but no systematic study of \ac{FS} or feature importance has been conducted for \acp{TM}. This work fills that gap.

\section{Overview of Tsetlin Machines}
\section{A Brief Overview of Tsetlin Machines}
\label{sec:background_tm}

\begin{figure}[ht]
\centering
\includegraphics[width=1\linewidth]{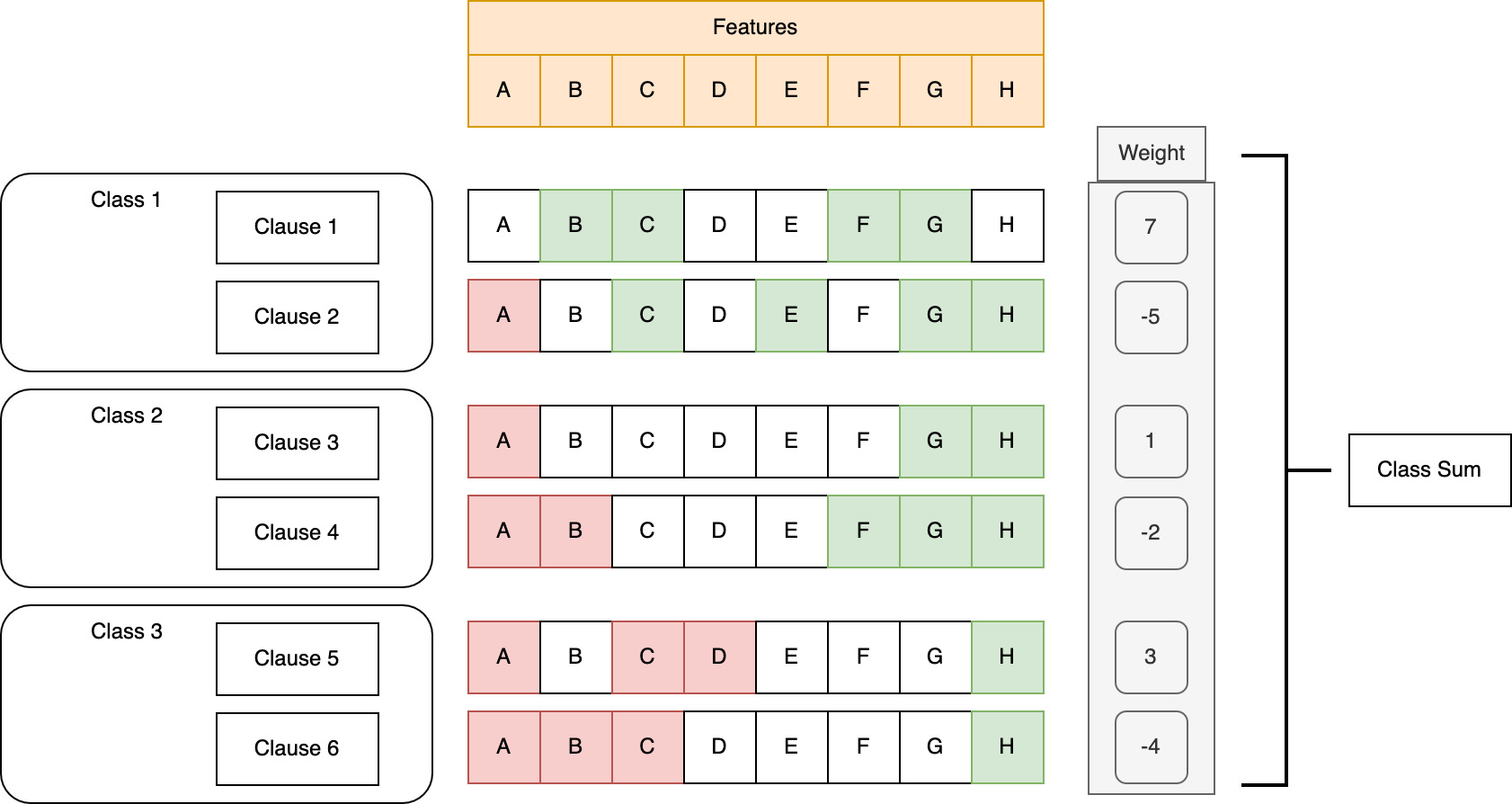}
\caption{Tsetlin Machine architecture: features (columns), clauses (rows), and class assignments.  Green literals are included, red literals are excluded, white are irrelevant.}
\label{fig:tm_arch}
\end{figure}

In short, the \ac{TM} \cite{Granmo2020_TM} is a logic-based pattern recognizer that learns human-readable propositional rules from binary inputs.  It is built upon two key components: \emph{clauses}, which are conjunctions of input \emph{literals}, and \acp{TA}, which decide whether to include or exclude each literal in each clause.  Learning proceeds via stochastic reinforcement on each automaton, using two types of feedback (detailed below).  Figure~\ref{fig:tm_arch} illustrates the overall architecture.

Given a binary input 
\[
  \mathbf{x}=(x_1,\dots,x_d)\in\{0,1\}^d,
\] 
each feature \(x_i\) is a literal \(l_i\equiv x_i\) and has a negation \(\neg l_i\equiv(1-x_i)\).  A \emph{clause} 
\[
  C_\ell(\mathbf{x})\;=\;\bigwedge_{i\in S_\ell} \bigl(l_i\bigr)^{\delta_i}
\;\in\{0,1\}
\]
is a conjunction over some subset \(S_\ell\subseteq\{1,\dots,d\}\) of literals.  For example,
\[
  C_\ell(\mathbf{x})
  = l_1\land\neg l_3\land l_7
  \;\Longleftrightarrow\;
  (x_1=1)\wedge(x_3=0)\wedge(x_7=1).
\]

Each clause \(\ell=1,\dots,M\) is assigned to a class \(c_\ell\in\{1,\dots,C\}\) and carries a polarity \(p_\ell\in\{+1,-1\}\), so that when the clause \emph{fires} (\(C_\ell(\mathbf{x})=1\)), it casts a vote with a weight \(w_\ell\).
\[
  w_\ell\,C_\ell(\mathbf{x}),
  \quad\text{where}\quad
  w_\ell \in\ R .
\]
The \emph{class sum} for class \(c\) aggregates these votes:
\[
  s_c(\mathbf{x})
  \;=\;
  \sum_{\ell=1}^M p_\ell\,C_\ell(\mathbf{x})\,\mathbf{1}\{c_\ell=c\}.
\]
The final prediction is 
\(\displaystyle \hat c = \arg\max_{c=1,\dots,C} s_c(\mathbf{x}),\)
.

Each literal–clause pair is governed by a Tsetlin Automaton, a two-action finite state machine that learns via stochastic Type I and Type II feedback \cite{tsetlin1973}:
\begin{itemize}
  \item \textbf{Type I feedback} rewards clauses that correctly fire on positive examples, reinforcing inclusion of literals that keep the clause true.
  \item \textbf{Type II feedback} penalizes clauses that erroneously fire on negative examples, reinforcing inclusion of literals that make the clause false.
\end{itemize}
Over many epochs, each automaton converges to either \emph{include} or \emph{exclude} its literal, thereby shaping each clause to capture relevant patterns.

Key hyperparameters are:
\begin{itemize}
  \item the total number of clauses \(M\) (often split equally per class and polarity),
  \item the \emph{threshold} \(T\), which controls the rate of feedback,
  \item the \emph{granularity} \(s\), which tunes how narrowly clauses match input patterns.
\end{itemize}
Despite their logical form, TMs scale to high-dimensional data (e.g., text \(n\)-grams, thermometer-encoded continuous features) by distributing learning across many simple automata.

Because each clause is a human-readable conjunction of literals, TMs offer innate rule-based explanations.  However, large datasets typically yield hundreds or thousands of clauses, making manual inspection intractable.  \ac{FS}—identifying which original features appear most frequently or carry the greatest weight across clauses—thus provides a natural way to distill and visualize the key signals driving TM decisions.

\section{Feature Selection Methods}

Table~\ref{tab:fs_methods_overview} summarizes all \ac{FS} methods evaluated: classical filters, post-hoc/wrapper explainers, and TM‐internal embedded scorers.

\begin{table*}[!htbp]
\centering
\caption{Overview of adapted and proposed feature‐scoring methods, their origins, and key references.}
\label{tab:fs_methods_overview}
\begin{tabular}{p{3.2cm}p{8.8cm}}
\toprule
\textbf{Method} & \textbf{Description / References} \\
\midrule
\multicolumn{2}{l}{\textbf{Filter Methods}} \\
MutualInfo & Measures mutual dependence between feature \(f\) and label. Based on Battiti (1994) \cite{Battiti1994_MI} and Brown et al.\ \cite{Brown2012_FS}. \\
Chi2       & \(\chi^2\) test of independence between \(f\) and label. Based on Forman (2003) \cite{Breiman2001_RF}. \\
Variance   & Removes features with low variance (baseline filter). \\
Random     & Assigns random scores (sanity check). \\
\midrule
\multicolumn{2}{l}{\textbf{Embedded (TM-Internal) Methods}} \\
Relevance        & Class-weighted TA engagement frequency: \(\sum_{\ell} \alpha_c\,\mathbf{1}\{\ell\text{ selects }f\}\). Granmo (2020) \cite{Granmo2020_TM}. \\
TM-Weight        & \(s_f = \max_{c,\ell}|w_\ell|\) for those literals \(\ell\) involving feature \(f\). Proposed here. \\
CW-Sum           & \(s_f = \sum_{c=1}^C \alpha_c\,|W_{c,f}|\), where \(\alpha_c\) is class-weight. Proposed here. \\
Support-CW-Sum   & As CW-Sum but \(\epsilon_c=1-\alpha_c\) replaces \(\alpha_c\). Proposed here. \\
CW-Feat          & Let \(\beta_{c,f} = |W_{c,f}| \big/ \sum_{c'}|W_{c',f}|\). Then
                   \(s_f = \sum_{c=1}^C \beta_{c,f}\,|W_{c,f}|\). Proposed here. \\
Margin           & Sort \(\{|W_{c,f}|\}_c\) as \(a_{(1)}\ge a_{(2)}\ge\cdots\); set \(s_f=a_{(1)}-a_{(2)}\). Proposed here. \\
Entropy          & Inverted Shannon entropy of \(\{\,|W_{c,f}|\,\}_c\). Based on Shannon (1948). \\
Gini             & Gini index on \(\{\,|W_{c,f}|\,\}_c\). Based on Gini (1912). \\
Stability        & Ratio of mean to std.\ of clause-weight history across epochs. Nogueira \& Brown (2017) \cite{Nogueira2017_Stability}. \\
TaylorCrit       & First-order change in true-class sum when flipping a literal. Inspired by Montavon et al.\ (2018) \cite{Montavon2018_LRP}. \\
VarDropout       & Sensitivity under random binary feature masks. Srivastava et al.\ (2014) \cite{Srivastava2014_Dropout}. \\
AblationImpact   & Impact of permanently ablating each feature. Zeiler \& Fergus (2014) \cite{Zeiler2014_Occlusion}. \\
SmoothStabil     & Sensitivity to small random input perturbations. Smilkov et al.\ (2017) \cite{Smilkov2017_SmoothGrad}. \\
\midrule
\multicolumn{2}{l}{\textbf{Wrapper Methods}} \\
Dropout         & Mask-one-feature sensitivity (leave-one-out). Koh \& Liang (2017) \cite{Koh2017_InfluenceFuncs}. \\
PermImportance  & Permutation importance of each \(f\). Fisher et al.\ (2018) \cite{Fisher2018_PermutationImportance}. \\
\midrule
\multicolumn{2}{l}{\textbf{Attribution / Ensemble NN-Inspired Methods}} \\
SHAP            & Shapley-value based contributions \(\phi_{f}\). Lundberg \& Lee (2017) \cite{Lundberg2017_SHAP}. \\
LIME            & Local surrogate linear approximation. Ribeiro et al.\ (2016) \cite{Ribeiro2016_LIME}. \\
IG              & Integrated Gradients attribution. Sundararajan et al.\ (2017) \cite{Sundararajan2017_IG}. \\
SmoothGradSq    & Squared-average of gradients over noisy inputs. Smilkov et al.\ (2017) \cite{Smilkov2017_SmoothGrad}. \\
VarGrad         & Variance of gradients over noisy inputs. Smilkov et al.\ (2017) \cite{Smilkov2017_SmoothGrad}. \\
\bottomrule
\end{tabular}
\end{table*}

We directly apply SHAP, LIME, IG, SmoothGrad, and Permutation Importance to the trained TM.
We further define indicator functions
\[
I_{\ell,f} \;=\;\mathbf{1}\{\text{clause }\ell\text{ contains the literal }l_f\}, 
\]
Then for each class \(c\) and feature \(f\) we accumulate
\[
  W^+_{c,f} 
  = \sum_{\ell=1}^M w_\ell\,I^+_{\ell,f}\,\mathbf{1}\{c_\ell=c\},
  \qquad
  W^-_{c,f} 
  = \sum_{\ell=1}^M w_\ell\,I^-_{\ell,f}\,\mathbf{1}\{c_\ell=c\}.
\]
We set
\[
  \mathrm{netW}_{c,f}
  = W^+_{c,f} - W^-_{c,f},
  \quad
  \mathrm{absW}_{c,f}
  = \bigl|\mathrm{netW}_{c,f}\bigr|.
\]
For the PosNeg variants we define
\[
  \mathrm{sumW}_{c,f}
  = W^+_{c,f} + W^-_{c,f},
  \quad
  \mathrm{absSumW}_{c,f}
  = \bigl|\mathrm{sumW}_{c,f}\bigr|.
\]

\section{Experimental Setup}

Table~\ref{tab:datasets_performance} summarizes dataset properties, and baseline TM accuracy/F1. Continuous features are binned into 10 thermometer levels. Splits were set to 60/20/20 if no standard split is provided.

\begin{table*}[!htbp]
\caption{Summary of datasets, TM parameters, and performance with all features at 30 epochs, 500 clauses, 10 thermometer bins per feature; parameters \emph{s} and \emph{T} found with 100 Optuna trials (s [0.9 to 20.0], T \{50,200,300,500,800\})}
\label{tab:datasets_performance}
\centering
\begin{tabular}{lrrrlrrrr}
\toprule
\textbf{Dataset} 
  & \textbf{\#Sam.} & \textbf{\#Feat.} & \textbf{\#Cls} 
  & \textbf{Source} 
  & \(\mathbf{s}\) & \(\mathbf{T}\) 
  & \textbf{Acc (\%)} & \textbf{F\(_1\)(\%)} \\
\midrule
Hierar.\ Bool.           & 500   & 20  & 2 & generated                  &  3.00 & 600 & 68.00 & 64.84 \\
Parity                   & 500   & 20  & 2 & generated                  &  3.00 & 600 & 44.00 & 42.53 \\
Feature Interact.        & 500   & 20  & 2 & generated                  &  3.00 & 600 & 51.00 & 50.76 \\
Balance Scale            & 625   & 4   & 3 & UCI                        &  3.00 & 600 & 88.00 & 61.12 \\
Banknote                 & 1 372 & 4   & 2 & UCI                        &  3.00 & 600 & 97.09 & 97.07 \\
Breast Cancer            & 569   & 30  & 2 & sklearn                    & 18.33 & 500 & 94.74 & 94.35 \\
Digits                   & 1 797 & 64  & 10& sklearn                    &  6.92 &  50 & 96.67 & 96.64 \\
Ecoli                    & 336   & 7   & 8 & UCI                        &  3.00 & 600 & 85.29 & 70.77 \\
Glass                    & 214   & 9   & 6 & UCI                        & 13.10 &  50 & 62.79 & 43.00 \\
Heart Disease            & 303   & 13  & 2 & OpenML                     &  3.32 & 300 & 81.48 & 81.38 \\
Ionosphere               & 351   & 34  & 2 & OpenML                     & 19.82 & 200 & 91.55 & 90.90 \\
Iris                     & 150   & 4   & 3 & sklearn                    & 14.80 & 300 & 90.00 & 90.15 \\
Pima Diabetes            & 768   & 8   & 2 & OpenML                     &  8.63 &  50 & 72.73 & 67.45 \\
Sonar                    & 208   & 60  & 2 & OpenML                     &  3.30 &  50 & 88.10 & 87.92 \\
Spambase                 & 4 601 & 57  & 2 & UCI                        &  3.00 & 600 & 89.90 & 89.31 \\
Steel Plates Faults      & 1 941 & 27  & 7 & OpenML                     &  8.34 &  50 & 78.92 & 74.16 \\
Transfusion              & 748   & 4   & 2 & UCI                        &  3.00 & 600 & 77.33 & 56.35 \\
Vehicle (Statlog)        & 846   & 18  & 4 & OpenML                     &  1.17 &  50 & 75.88 & 75.13 \\
Wine                     & 178   & 13  & 3 & OpenML                     &  9.70 & 800 & 94.44 & 94.53 \\
\bottomrule
\end{tabular}
\end{table*}

TM models were implemented in Python 3.12.4 on Mac M1 Pro 16 GB RAM. We fix 500 clauses, 10 bins per feature, and tune $(s,T)$ via Optuna (100 trials). Epochs=30 for score history. See Table~\ref{tab:datasets_performance} for found hyperparameters. Each Accuracy measure taken is an average over 10 trials.

We assess FS rankings by these evaluation protocols:
\begin{itemize}
  \item \textbf{Insertion/Deletion}: mask/unmask top-$k$ features
  \item \textbf{\ac{ROAR}}: remove top-$k$ features
  \item \textbf{\ac{ROAD}}: replace top-$k$ with marginal samples
\end{itemize}
\ac{ROAD} should be the most advanced evaluation protocol, however, others are included for thorough comparison. Each protocol creates several subsets of features given by the chosen \ac{FS} technique. These subsets are fed into 10 new \acp{TM}, with the same parameters as the original where we took the ranking from. Their accuracy is then averaged and plotted against the number of features \emph{K}.

\begin{figure}[!htbp]
\centering
\includegraphics[width=1\linewidth]{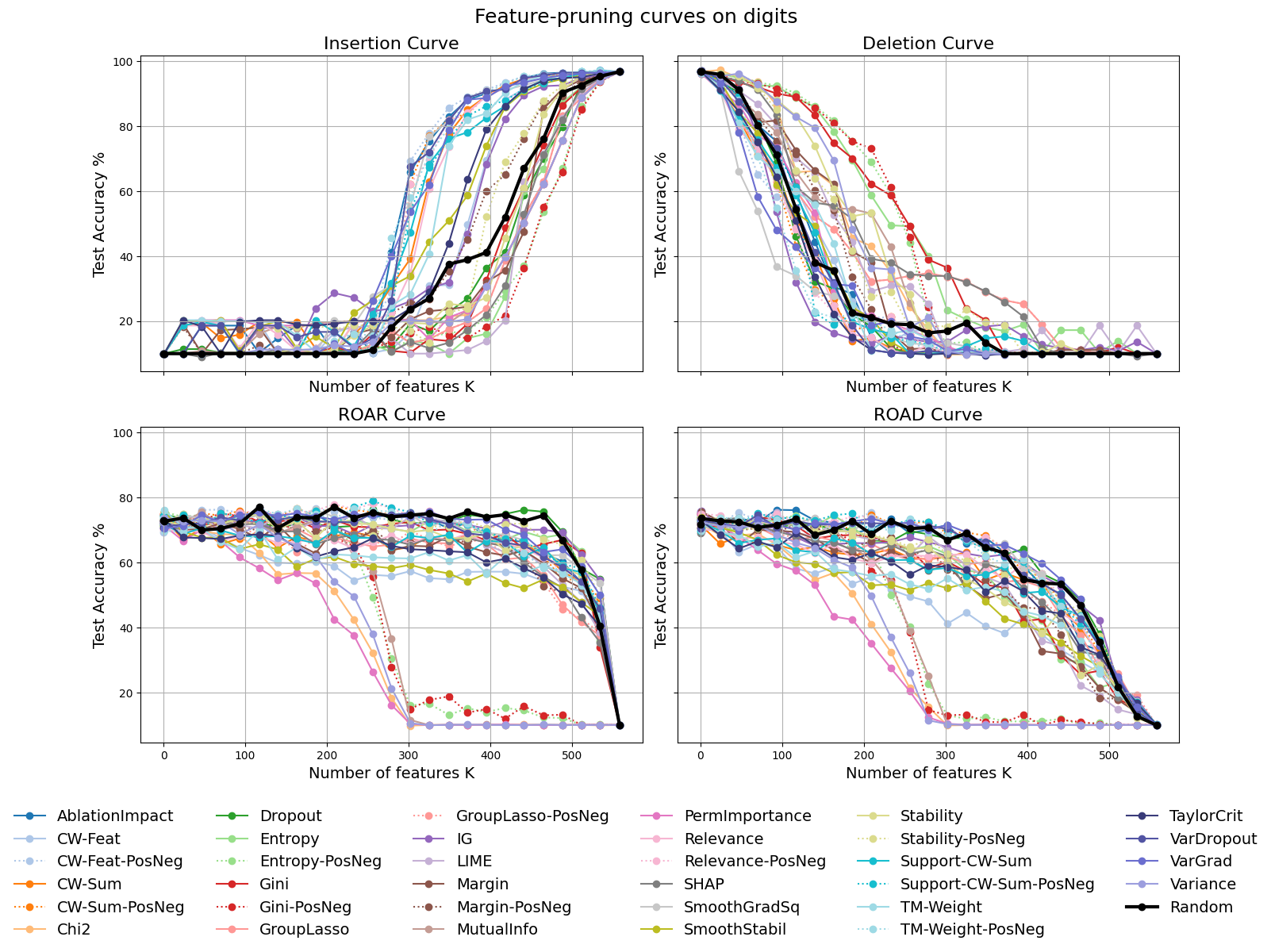}
\caption{Example of the pruning curves of the used Evaluation Protocols on the Digits dataset. Black is random as a reference.}
\label{fig:curvesDig}
\end{figure}

\begin{figure}[!htbp]
\centering
\includegraphics[width=1\linewidth]{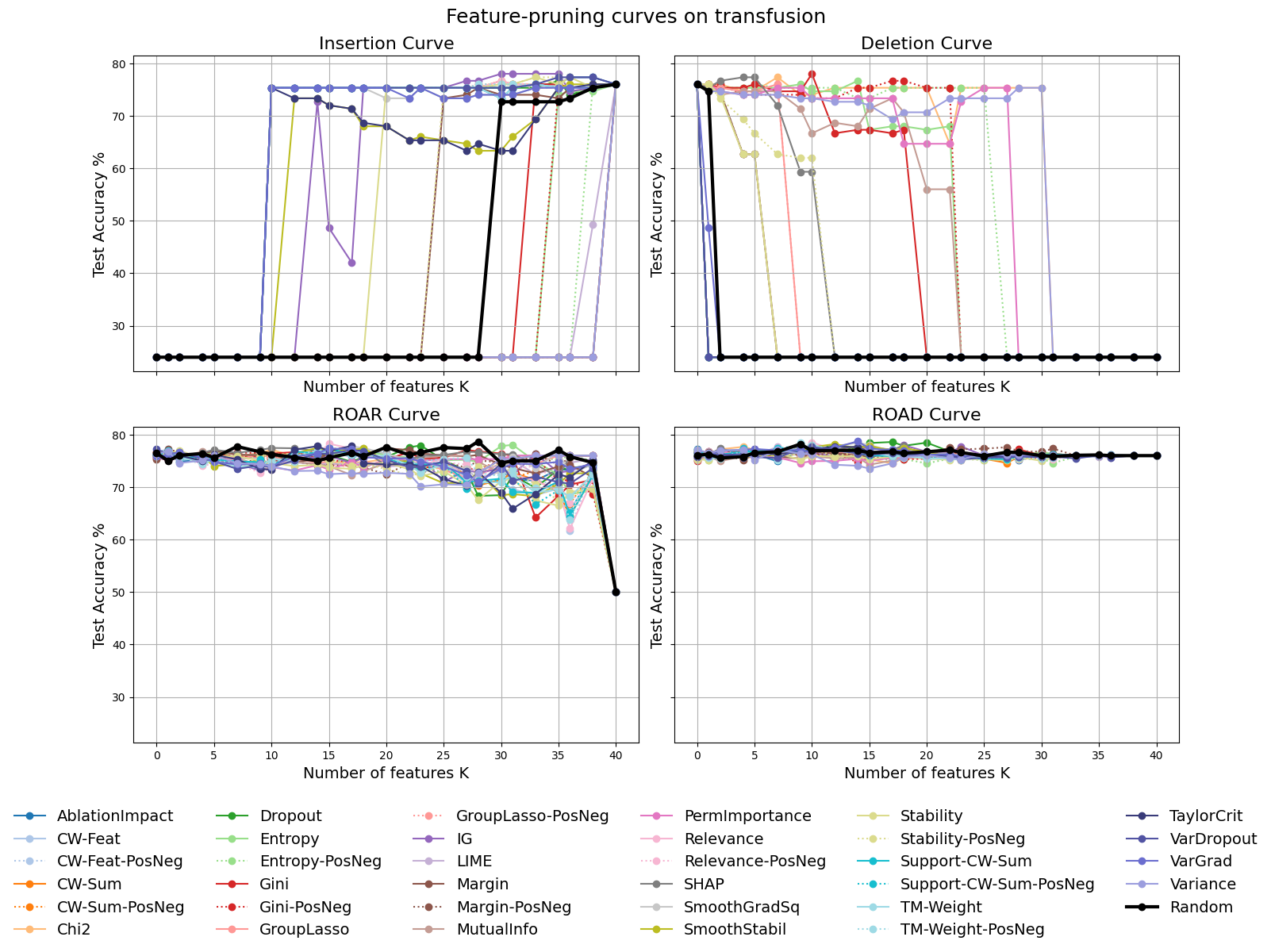}
\caption{Example of the pruning curves of the used Evaluation Protocols on the Transfusion dataset. Black is random as a reference.}
\label{fig:curvesTrans}
\end{figure}

\newpage
Figure \ref{fig:curvesDig} and \ref{fig:curvesTrans} illustrates an example of these curves. Comparison of those two figures shows how the spread of assigned importance by a different \ac{FS} correlates with the natural intuition. As in, Digits is has many redundant features, and medical datasets tend to have a single telling pattern, that once is found, defines the label. Moreover, Digits contains continuous features, and therefore thermometer encoding was used, this increased the redundancy.

\newpage

\ac{ROAD} curve of the Transfusion datasets seems nearly constant for all \ac{FS} techniques, which could be caused by a much lower achieved F1 score by the used model, than in the Digits case. Achieved F1 score was 56 for Transfusion and 96 for Digits.

\section{Results}

\begin{figure}[!htbp]
\centering
\includegraphics[width=1\linewidth]{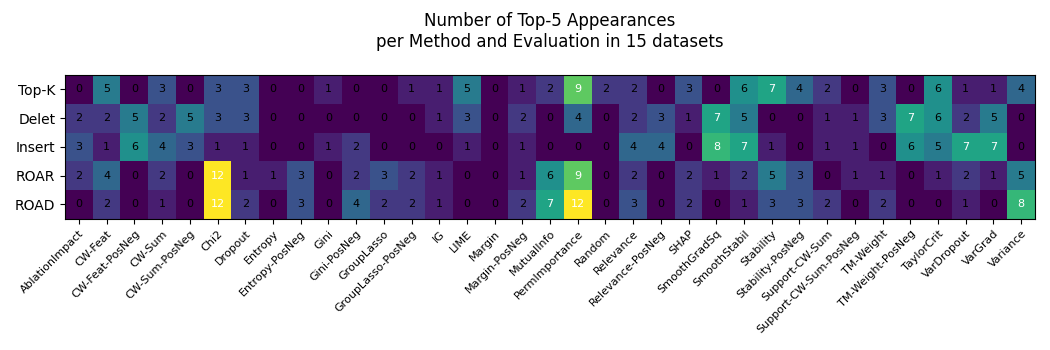}
\caption{Appearances in top-5 \ac{AUC} scores across datasets and protocols. We look for consistency among all, or the most advanced Evaluation protocol only - the \ac{ROAD}.}
\label{fig:fs_top}
\end{figure}

Table in Figure~\ref{fig:fs_top} shows the count of each method appearing in the top-5 per protocol.

Chi2, Variance and MutualInfo filter methods, seem to dominate right after PermImportance wrapper. This dominance of simple filters can be attributed to our thermometer‐encoding, which expands each original variable into several binary features; marginal‐based statistics (variance, X², MI) then readily identify those with the greatest spread or class‐association, without needing to model inter‐feature interactions. Best among its category are Chi2, PermImportance and Stability-PosNeg.

\newpage

\begin{figure}[!htbp]
\centering
\includegraphics[width=0.9\linewidth]{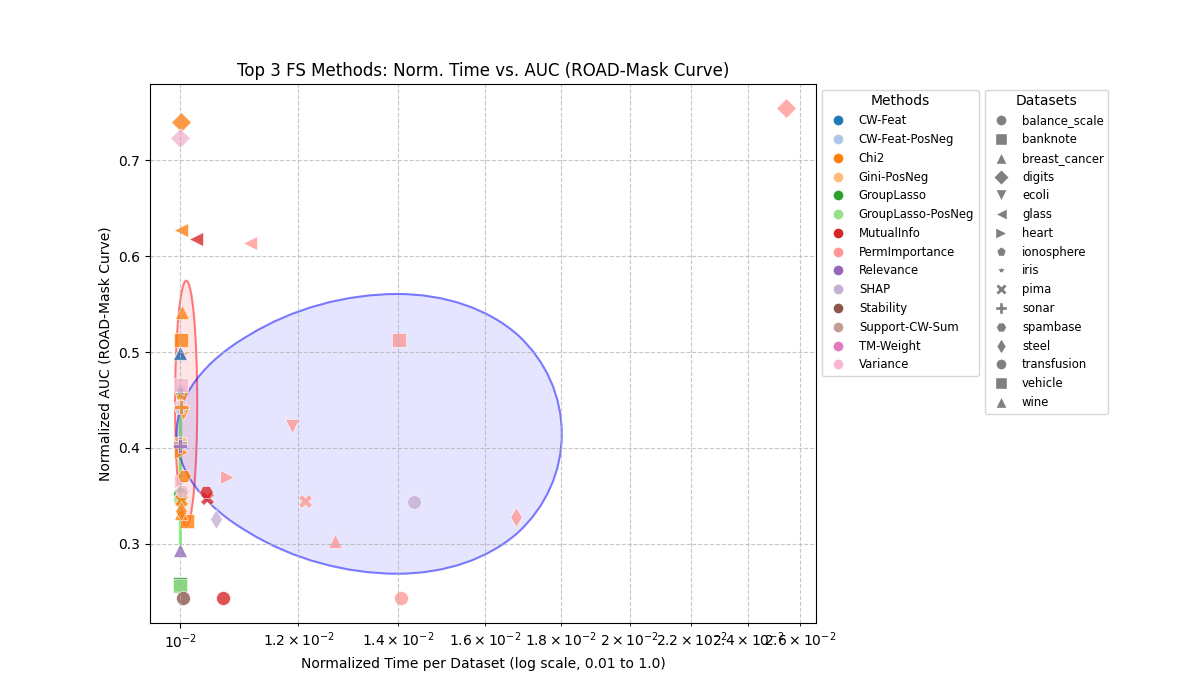}
\caption{Normalized \ac{AUC} vs. ranking time for top-3 methods per dataset. Grouped average of each category by colour Red is Filter, Green is Embedded, Blue is Wrapper}
\label{fig:fs_tradeoff}
\end{figure}

Figure~\ref{fig:fs_tradeoff} plots per-dataset top-3 methods' \ac{AUC} vs. normalized ranking time. It also shows efficiency of each category vs its \ac{AUC} with \ac{ROAD} evaluation. Here we see that filters (red) are extremely fast but plateau at moderate \ac{AUC}, whereas wrappers (blue) achieve slightly higher \ac{AUC} at much greater cost—reflecting their retraining or perturbation loops—and embedded methods (green) strike a middle ground by leveraging TM internals at modest overhead.

\begin{figure}[!htbp]
\centering
\includegraphics[width=1\linewidth]{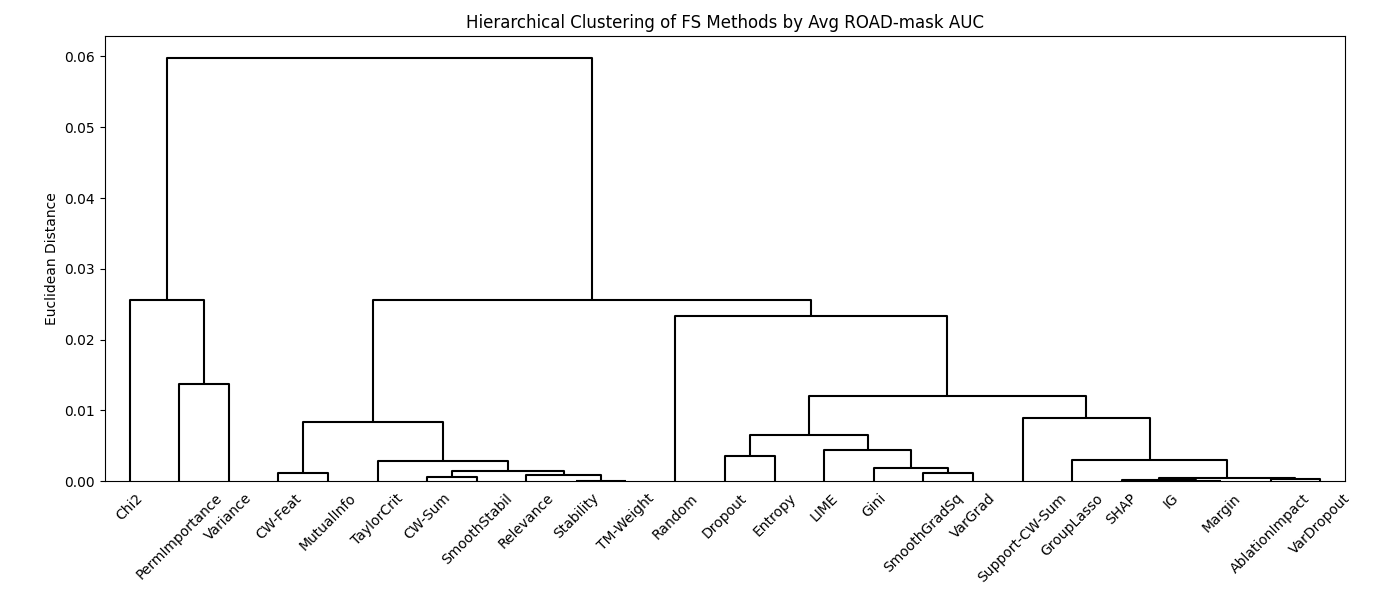}
\caption{Dendrogram of the used FS techniques, averaged among all datasets}
\label{fig:dendrogram}
\end{figure}

\newpage
We used pairwise Euclidean distances between each method’s average ROAD-mask AUC profiles across all datasets, and average-linkage clustering to produce the dendrogram in Figure~\ref{fig:dendrogram}. A few noteworthy patterns emerge:
\begin{itemize}
    \item Filter methods (MutualInfo, Chi2, Variance) form a cleanly separated branch, reflecting their shared reliance on simple, data-centric statistics.
    \item LIME and SHAP do not cluster together; instead each associates with a different subset of embedded scorers, hinting at their distinct attribution biases even though both are “post-hoc” wrappers, which suggests that SHAP’s global Shapley‐based attributions and LIME’s local surrogate fits emphasize different aspects of TM clause outputs.
    \item The two TM-internal scorers CW-Feat and CW-Sum sit in their own tight subcluster, underscoring their similar nature.
    \item Proposed embedded methods occupy a middle ground—structurally closer to each other than to filters or neural wrappers—thereby highlighting the unique, clause-based nature of TM feature selection.

\end{itemize}

Together, these clusters confirm that methods group by their reliance on marginal versus interaction‐aware versus clause‐based signals.

\begin{figure}[!htbp]
\centering
\includegraphics[width=0.9\linewidth]{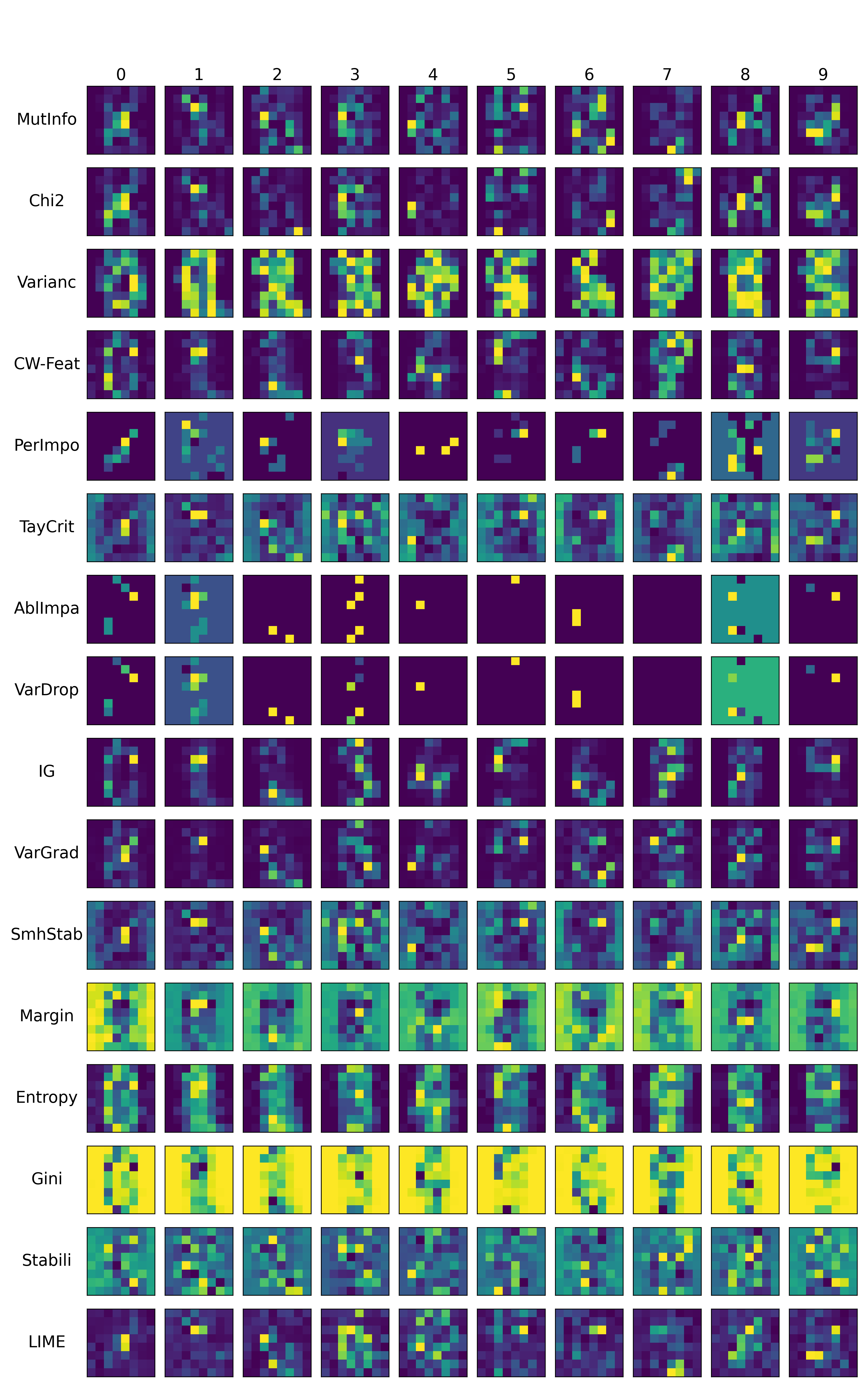}
\caption{Plots of the important features according to different \ac{FS} Techniques - Proposed \ac{TM} Embedded \ac{FS} techniques are the most interpretable}
\label{fig:digits}
\end{figure}

Lastly, we validate alignment of the feature rankings with human intuition on the Digits dataset (Fig.~\ref{fig:digits}). 
\emph{Filters} Chi² and MutualInfo light up similar high-variance pixels at the digit boundaries, Variance focuses on the outline. \emph{Wrappers} PermanentImportance, IG LIME and VarGrad reveal pixels whose shuffling most degrades class sums, but are unreadable.\emph{Embedded} CW-Feat, TayCrit, show numbers; AlbImpa and VarDrop are heavily selective with their importance, and show individual pixels.
SmoothStability and Margin seem to produce similar results to Wrappers.
Entropy, Gini, Stability, show the numbers almost readably.

\newpage

\section{Discussion and Future Work}
\label{sec:discussion}
Chi2 filter method either dominates, or is in the top 5 rankings, according to \ac{ROAD}. It is on par with Permanent Importance wrapper, and Variance filter is lagging only one place behind them. Such high ranking of all the filters suggests that filters would be a great addition to the \ac{TM}. This is most likely due to the thermometer encoding of all input features.

Future work could leverage feature scores to highlight not only important inputs but also which clauses and rules they influence, enabling symbolic-level interpretability.

Our experiments show that simple \ac{TM}-embedded scorers (e.g.\ CW-Sum, \ac{TM}-Weight) often match or exceed the quality of \ac{NN}-inspired explainers at far lower cost.
Classical filters are fast but miss higher level interactions. Key insights:
\begin{itemize}
  \item Embedded \ac{TM} scorers effectively capture clause-based interactions.
  \item Wrappers remain costly on Boolean inputs.
  \item Hybrid two-stage \ac{FS} (filter and embedded) could further reduce cost.
  \item Online adaptive \ac{FS} during \ac{TM} training seems like a promising direction.
  \item \ac{FS} ranking may be used to help with explainability, once the number of clauses grows too large
\end{itemize}

We presented the first systematic benchmark of filter, wrapper/post-hoc, and embedded \ac{FS} methods on \acp{TM}, evaluated via Insertion/Deletion, \ac{ROAR}, and \ac{ROAD}. \ac{TM}-internal scorers offer a middle ground, speed–quality trade-off. Moreover they also carry on the \ac{TM}'s natural explainability, whilst other \ac{FS} method fail at showing a human understandable scores. Future work will explore hybrid and online \ac{FS} strategies, and compare directly against the same techniques on neural networks.

\newpage

\bibliographystyle{IEEEtran}
\bibliography{References}

\end{document}